\documentclass[letterpaper, 10 pt, conference]{ieeeconf}  

\IEEEoverridecommandlockouts                              

\usepackage{ragged2e}    
\usepackage{array}       
\newcolumntype{J}[1]{>{\justifying\arraybackslash}p{#1}}

\usepackage{amsmath} 
\usepackage{amssymb}  
\usepackage{graphicx}
\usepackage{soul}
\usepackage{xcolor}
\usepackage[acronym]{glossaries}
\usepackage[ruled,vlined,noend]{algorithm2e}
\usepackage{algpseudocode}%
\usepackage{multicol}
\usepackage{cite}
\usepackage{siunitx}
\usepackage{stfloats}
\usepackage{caption}
\usepackage{booktabs}
\usepackage{multirow}
\usepackage{siunitx}
\usepackage{ragged2e}   
\usepackage{array}      
\usepackage{graphicx}   
\usepackage{subcaption}
\newcolumntype{P}[1]{>{\justifying\arraybackslash}m{#1}}

\usepackage[compact]{titlesec}         
\titlespacing{\section}{0pt}{0pt}{0pt} 
\AtBeginDocument{
  \setlength\abovedisplayskip{0pt}
  \setlength\belowdisplayskip{0pt}}
  
\title{\LARGE \bf An Actionable Hierarchical Scene Representation Enhancing Autonomous Inspection Missions in Unknown Environments}

\author{Vignesh Kottayam Viswanathan, Mario A.V Saucedo, Sumeet Gajanan Satpute,\\ Christoforos Kanellakis and  George Nikolakopoulos
\thanks{The authors are with Robotics and AI, Luleå University of Technology 97187, Luleå,Sweden
        {\tt\small \{vigkot, marval, sumsat, chrkan and geonik\}@ltu.se}}%
\thanks{The authors would like to thank Elias Small and Jude Marroush for assisting in field evaluations.}
\thanks{This work has been partially funded by the European Union's Horizon Europe Research and Innovation Programme, under the Grant Agreement No.101138451 PERSEPHONE.}
}

\newacronym{lsg}{LSG}{Layered Semantic Graphs}
\begin{document}

\maketitle
\thispagestyle{empty}
\pagestyle{empty}

\begin{abstract}
In this article, we present the \textit{Layered Semantic Graphs} (LSG), a novel actionable hierarchical scene graph, fully integrated with a multi-modal mission planner, the FLIE: \textit{A First-Look based Inspection and Exploration planner}~\cite{viswanathan2023towards}. The novelty of this work stems from aiming to address the task of maintaining an intuitive and multi-resolution scene representation, while simultaneously offering a tractable foundation for planning and scene understanding during an ongoing inspection mission of apriori unknown targets-of-interest in an unknown environment. The proposed LSG scheme is composed of locally nested hierarchical graphs, at multiple layers of abstraction, with the abstract concepts grounded on the functionality of the integrated FLIE planner. Furthermore, LSG encapsulates real-time semantic segmentation models that offer extraction and localization of desired semantic elements within the hierarchical representation. This extends the capability of the inspection planner, which can then leverage LSG to make an informed decision to inspect a particular semantic of interest. We also emphasize the hierarchical and semantic path-planning capabilities of LSG, which could extend inspection missions by improving situational awareness for human operators in an unknown environment. The validity of the proposed scheme is proven through extensive evaluations of the proposed architecture in simulations, as well as experimental field deployments on a Boston Dynamics Spot quadruped robot in urban outdoor environment settings. 
\end{abstract}
\section{Introduction}
Incorporation of a semantic dimension to conventional geometric mapping has created a rich understanding of the environment when viewed from a robot's perspective. A relevant example of such a metric-semantic representation is the 3D Scene Graphs (3DSG)~\cite{kim20193}. While 3D scene graphs maintain expressive multi-resolution model of the environment, existing methods focus on exploiting such a representation for task domains such as robot manipulation~\cite{jiao2022sequential}, task planning~\cite{agia2022taskography}, localization and mapping~\cite{looper20233d} and semantic navigation~\cite{bavle2022situational}.

 
In this work, we consolidate the need for hierarchical 3D scene graphs by establishing a connection between high-level mission planning in unknown environments and real-time scene graph synthesis. We ground abstract concepts in the hierarchical representation based on instantaneous planner inputs, dynamically populating relevant graph layers depending on the planning mode (e.g., inspection or exploration). Instead of maintaining a global subgraph for each abstraction layer, we propose a nested scene graph scheme, called \gls{lsg}. A 3D Layered Semantic Graph consists of local hierarchical graphs for each semantic layer and node, with deeper local graphs nested within the attributes of nodes in higher layers. This ensures that the chosen path planner is exposed to selective localized graph representations at the necessary abstract layer during hierarchical path planning (see Section.~\ref{sec:hpp_spp}), while also maintaining an intuitive representation of the information captured during the mission.

In view of related works, we bifurcate the study into two aspects:~\textbf{i) Hierarchical representation of unknown environments:} The authors in~\cite{galindo2005multi} laid the foundation for representing a tele-operated robot's perceived environment as a spatial-semantic hierarchy. The authors bifurcated the representation into a metric-based hierarchy and one for symbolic abstract concepts, "anchoring" the ground abstract notions (e.g., \textit{bedroom}) to spatial nodes with 3D positional information. This concept has been further expanded in~\cite{zender2008conceptual}, by introducing a multi-layered metric-semantic approach, augmenting the graph with spatial, topological, and conceptual information to enable reasoning. A more recent work by~\cite{armeni20193d} introduced a 3D scene graph structure based on 3D mesh and RGB data, clustering nodes by semantic concepts, such as \textit{buildings}, \textit{rooms}, and \textit{objects}. Furthermore, the article~\cite{rosinol2021kimera} developed 3D dynamic scene graphs, generating the graph \textit{bottom-up} from low-level parametric data. This work was further expanded in~\cite{hughes2022hydra} to real-time graph construction with incremental 3D scene graph generation, creating rich metric-semantic maps. Other studies, such as~\cite{zhou2022hivg}, focused on multi-layered hierarchical models for indoor navigation, using visibility graphs and geometric extraction through graph partitioning algorithms. The work in~\cite{bavle2022situational} proposed a hierarchical representation integrating topological, semantic, and spatial concepts, using a \textit{bottom-up} process to build graphs from LiDAR data, with layers for localization, semantic abstraction, and segmented meshes.~\textbf{ii) Inspection planning in unknown environments}: The current state-of-art on inspection planning in unknown environments is presented with the focus on providing optimal coverage~\cite{song2021view,brogaard2021towards}. While satisfying the purpose of obtaining a 3D map of the environment in the end, such implementations are designed to exhaustively cover the entire operational environment. In~\cite{dharmadhikari2023semantics}, the authors proposed an indoor semantics-aware inspection scheme, leveraging volumetric exploration and subsequent geometric coverage of observed candidate semantics within the operating environment. Recently,~\cite{ginting2024semantic} presented a belief-behaviour graph based semantic inspection in indoor environments that addressed the need to gather semantic characteristics of an object of interest, based on modelling the decision for the optimal behaviour on scene understanding and uncertainty in observations.

\textbf{Contributions:} With respect to the existing state of the art works, the scientific contributions comprising the proposed architecture in this work is stated as follows:
Firstly, we present the foundational solution for the incremental and real-time construction of a planner-centric actionable hierarchical scene graph (in Section.~\ref{sec:3D_LSG}): \textbf{ 3D Layered Semantic Graphs} (LSG). \textit{LSG} is a \textit{layered} and \textit{nested} graph built \textit{top-down} to register and maintain characteristic actionable knowledge of an apriori unknown inspection target during the lifetime of the inspect-explore autonomy. The graph is \textit{layered} and \textit{nested}, such that the graphs at every layer is populated based on the abstraction representing different levels of information provided by the planner. The \textit{nested} concept is realized through assigning graphs at deeper layers as attributes of the parent node in the layer above. The nodes within each graph model semantic concepts with intra-layer edges capturing spatial or symbolic relationships. The graph is \textit{actionable}, as the constituent nodes, edges and its encoded attributes can be leveraged by the mission planner to address runtime objectives during the mission.  
    
Our second contribution (in Section.~\ref{sec:xflie}) is \textbf{xFLIE: Enhanced First-Look Inspect-Explore Autonomy}, the first multi-modal autonomy scheme to construct and exploit, in real-time, a 3D hierarchical scene representation for semantics-aware inspection of apriori unknown targets-of-interest in unknown environments. xFLIE is a modular autonomy comprising of an \textit{inspection} planner, an \textit{exploration} planner and the 3D \textit{LSG} constructor working in tandem to address mission objectives. It leverages the map-free reactive planning provided by the FLIE~\cite{viswanathan2023towards} autonomy and exploits the high-level semantic abstractions within LSG to make an informed decision to inspect a semantic of interest.

\section{3D Layered Semantic Graphs}\label{sec:3D_LSG}

\subsection{Preliminaries}\label{sec:prelims}

In this work, we construct a hierarchical scene graph composed of multiple layers of abstraction of an unknown environment. The abstract concepts are defined based on the functionality of the integrated autonomy, i.e inspection and exploration planner. We denote $\mathbb{L}|~\mathbb{L} = [\textit{Target},\textit{Level},\textit{Pose},\textit{Feature}]$ as the set of abstraction layers that constitute LSG. Let $\mathcal{G}_{l} = (V_l,E_l,A_l)$ be an undirected graph of layer $l \in \mathbb{L}$ with an ordered set of nodes $V_l$ and edges $E_l$. We denote $a_i \in A_l$ as the attribute of a node $v_i$ encoded with the necessary metric, sensor, and semantic information, in addition to the graph of the next layer.

The LSG in this work excludes the typical consideration of inter-layer edges during construction and subsequently for path-planning. Instead, we introduce the concept of a layer frontier node $v^H_l \subseteq V_l \in \mathcal{G}_l$, which acts as a bridge between local graphs in different layers. A layer frontier node functions as both a child in its local graph and a parent in the nested graph, exploiting the latent property of nested graphs during hierarchical planning. This means that local graphs remain hidden from the planner until required, allowing the planner to access and expand them, as and when needed, when navigating across layers.

Additionally, we subscribe to the following notations where the operator $||\cdot||$ denotes the Euclidean norm and the operator $|\cdot|$ denotes the cardinality of a set, which refers to the number of elements in the respective set. 

\textbf{Terminologies:} In this work, we distinguish between the terms, \textit{semantic targets} and \textit{semantic features}, as follows: (a) \textit{semantic targets} refer to targets of interest, such as \textit{car}, \textit{house}, or \textit{truck}, that is observed in the environment as candidates for further inspection; and (b) \textit{semantic features} refer to the observed and segmented features of a semantic target, such as the \textit{window} or \textit{door} of a house, or the \textit{hood} of a car, which are the focus of the inspection mission.

\subsection{Layers Construction}\label{sec:layer_construction}

\begin{figure}[htbp]
    \centering
    \includegraphics[width = 0.8\linewidth]{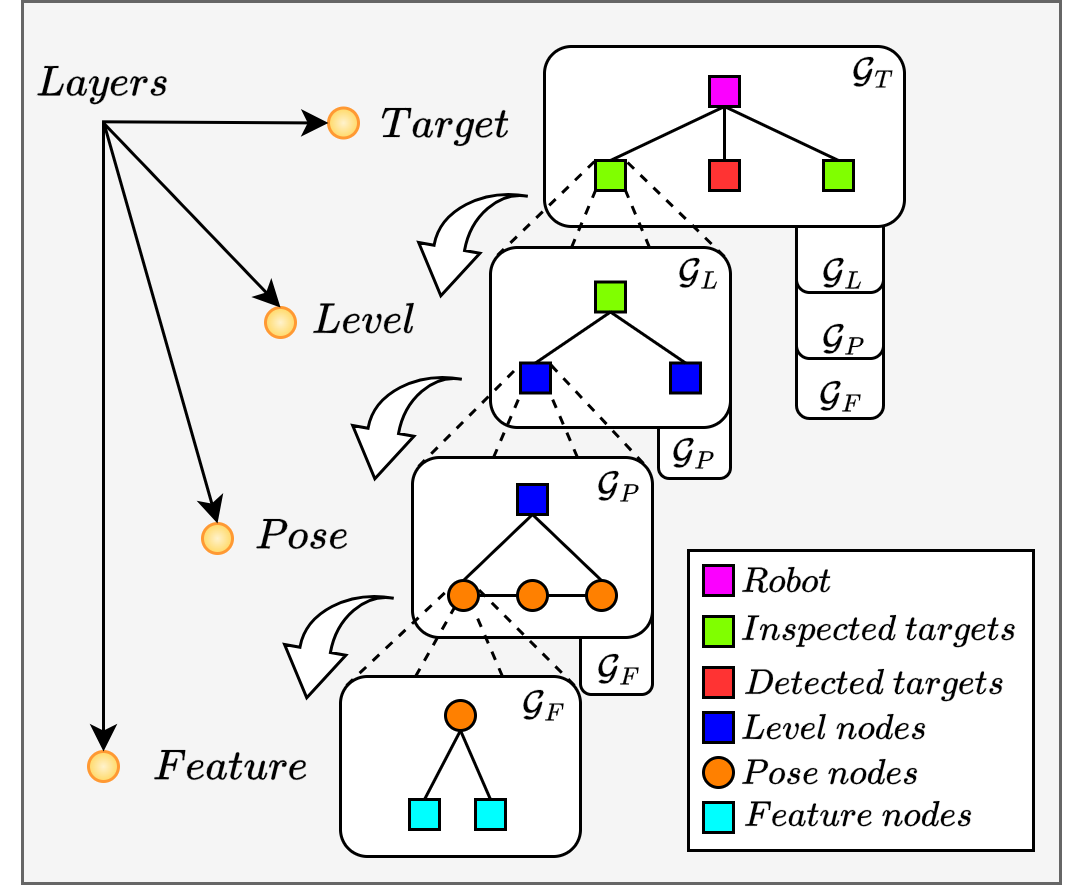}
    \caption{A visual insight into the nesting structure of multiple abstraction layers that comprises 3D LSG. The top-most layer, i.e the $\textit{Target}$ layer maintains $\mathcal{G}_T$ graph representation of detected and inspected semantic targets. For each inspected target, lower layers abstraction, i.e $\textit{Level},~\textit{Pose},~\textit{Feature}$ layers maintain their corresponding local graph representation $\mathcal{G}_L,~\mathcal{G}_P,~\mathcal{G}_F$ respectively. The \textit{jump-out arrow} beside each layer indicates the concept of nesting the lower graph representation within the node in the parent layer. The collapsed \textit{tabs} (shown under the parent layers) with the graph markings showcase the protractible nature of 3D LSG that is leveraged during planning.}
    \label{fig:LSG_structure}
\end{figure}

Figure~\ref{fig:LSG_structure} presents a detailed insight into the internal composition of 3D Layered Semantic Graphs constructed during the mission.

\textbf{Target Layer:} The $\textit{Target}$ layer maintains the $\mathcal{G}_T$ graph, which comprises of child nodes ($v_T \in V_T$), representing semantic targets, i.e specific semantics-of-interest for inspection (e.g \textit{house}, \textit{car}, or \textit{truck}). The parent node within the $Target$ layer is modelled to be a representation of the robotic agent, encoded with information on the current 6D pose of the robot $\textbf{X}_{odom} \in \mathbb{R}^3 \times \mathbb{SO}(3)$. The child nodes are differentiated into two main subsets $\{v^I_T\},\{v^D_T\} \subseteq V_T$: (a) $v^D \in \{v^D_T\}$, representing the detected semantic targets registered during exploration, and (b) $v^I \in \{v^I_T\}$, representing the inspected semantic targets.

During initial registration, $v^D$ is encoded with attributes such as: (a) estimated 3D position $\hat{\textbf{X}}^D_{T} \in \mathbb{R}^3$, (b) the corresponding RGB image, (c) the segmentation characteristics, i.e. the semantic class, detected confidence and segmented mask area, and (d) the utility of the detected node. During inspection, the attribute of the node is augmented with the nested graph information of the $\textit{Level}$ layer along with the bounding polygon $\mathbb{P}$ composed of the individual view-pose configurations maintained within the first level of inspection. Since FLIE and LSG inherently do not construct and maintain a global 3D map representation of the environment, polygon-based containment eliminates and prevents duplicate registration of semantic targets during exploration.

Intra-layer symbolic connections are established between the parent node and the registered child nodes. Moreover, any two $v^I$ share an edge if the condition for $\textit{Target}$ layer traversal is satisfied. This condition checks, during exploration, whether there exists a visually unobstructed region
between the current inspection semantic target ($v^{I,curr}_T$) and the ones previously observed and within its Field-Of-View (FOV), $v^{I,k}_T \in \{v^I_T\}| k = [1,2,..,|\{v^I_T\}|],v^{I,k}_T \neq v^{I,curr}_T$. This distills into a Point-In-Polygon (PIP) check, where if $\hat{\textbf{X}}^D_{T}$ of $v^{I,k}_T$ lies within $\mathbb{P}^{I,k}_T$, they would share a weighted edge that can be used for path planning. 

\textbf{Level Layer:} The $\textit{Level}$ layer graph $\mathcal{G}_L$ of $v^{I,curr}_T$ is initialized and populated with $v_L \in V_L$ by the planner during inspection and represents the levels of inspection carried out for $v^{I,curr}_T$. Under the FLIE autonomy~\cite{viswanathan2023towards}, a level of inspection is set to be completed if the planner revisits and recognizes a previously inspected surface for a given level of inspection. The readers are directed to the work presented in~\cite{viswanathan2023towards} for an in-depth understanding of the FLIE autonomy framework. 

During registration, $v_L$ is encoded with the following attributes: (a) 3D position of the level, and (b) the nested $\textit{Pose}$ layer graph $\mathcal{G}_P$ of the current level of inspection ($v^{curr}_L$). Within $\mathcal{G}_L$, $v^{I,curr}_T$ serves as the parent node sharing a symbolic edge with the child-level nodes. However, two adjacent $v_L$ nodes share a weighted edge corresponding to the Euclidean distance between their respective 3D spatial coordinates.

\textbf{Pose Layer:} Once $v^{curr}_L$ is initialized with the nested $\textit{Pose}$ layer graph $\mathcal{G}_P$, the planner populates the graph with nodes $v_P \in V_p$ corresponding to the commanded inspection view-pose configurations. Each $v_P$ is encoded with the attributes such as: (a) a 6D pose of the view configuration, (b) a RGB image captured at $v_P$, and (c) the nested $\textit{Feature}$ layer graph $\mathcal{G}_F$ consisting of the observed semantic features at the current pose node ($v^{curr}_P$). 

Intra-layer weighted connections exists between the parent node, i.e $v^{curr}_L$ and the initial and terminal child $v_P$ nodes. Furthermore, each $v_P$ shares a weighted edge with its adjacent nodes based on the Euclidean distance of its corresponding 3D location.

\textbf{Feature Layer:} The $\textit{Feature}$ layer comprises of relevant semantic features, i.e \textit{windows}, \textit{doors}, \textit{hood} or \textit{tailgate}, registered as nodes ($v_F$) within $\mathcal{G}_F$, observed at the $v^{curr}_P$. The nodes are encoded with the following attributes: (a) 3D position, and (b) segmentation characteristics, i.e semantic class, detected confidence and segmented area. With respect to intra-layer edges, each $v_F$ shares a symbolic edge with the parent pose node $v^{curr}_P$ from which it was observed.

\section{xFLIE: Enhancing Inspection in Unknown Environments with Actionable Hierarchical Scene Representations}\label{sec:xflie}

\begin{figure}[htbp]
    \centering
    \includegraphics[width = \linewidth]{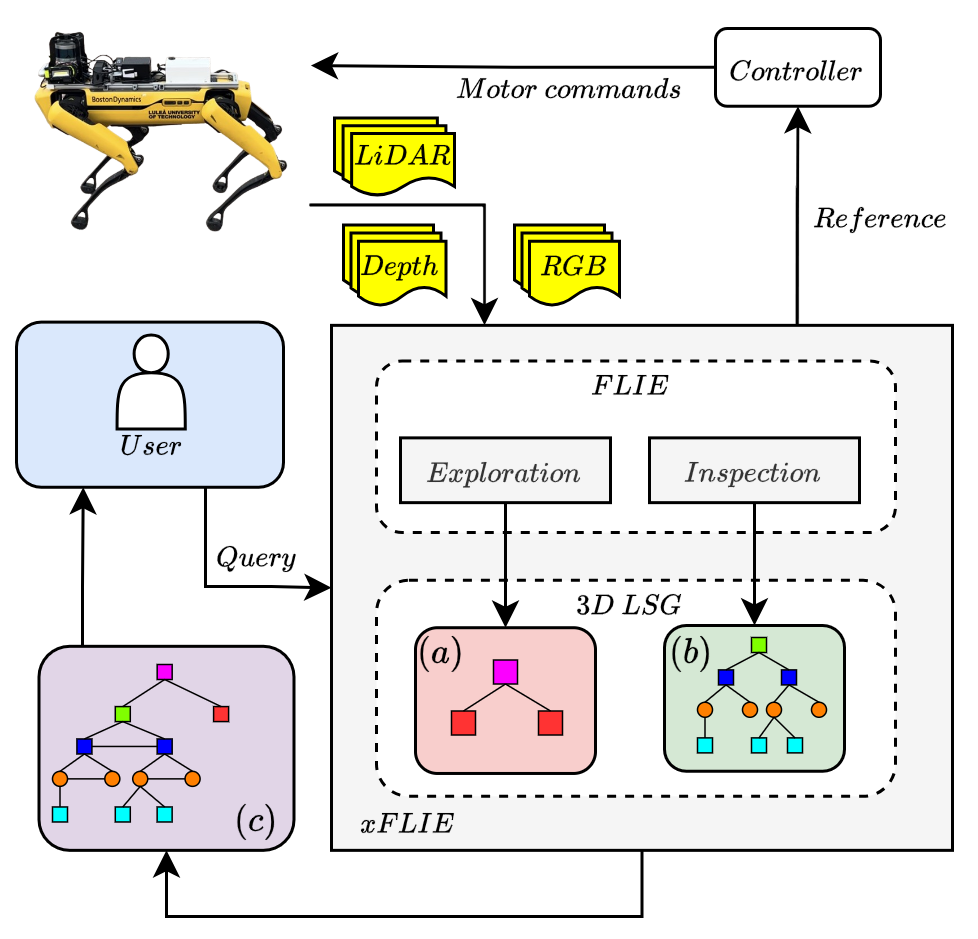}
    \caption{A visual overview of the xFLIE architecture composed of FLIE planner, an inspect-explore autonomy~\cite{viswanathan2023towards}, and the 3D LSG. xFLIE uses RGB, Depth and LiDAR measurements (shown on \textit{top-center}) to segment and localize desired semantics during inspection and exploration. The FLIE planner populates the corresponding abstract layers within LSG, where (a) the $\textit{Target}$ layer is populated during exploration and in (b) the $\textit{Level},~\textit{Pose},~\textit{Feature}$ layers are populated during inspection. The output of the xFLIE architecture is (c) an actionable hierarchical scene graph which can then be used by a human-operator to share semantic queries to the onboard autonomy.}
    \label{fig:lsg_pipeline}
\end{figure}

Figure.~\ref{fig:lsg_pipeline} presents the integrated multi-modal FLIE autonomy, composed of an \textit{inspection} and an \textit{exploration} planner, with the 3D LSG constructor for autonomous inspection missions. In general, the autonomy follows an {explore}-{inspect}-{explore} approach until there are no remaining semantic targets in the observable environment to inspect. During \textit{exploration}, the planner executes two main policies: (a) $\pi^{expl:360}$, an initial $360^{\circ}$ survey upon mission initialization, and (b) $\pi^{expl:LE}$, a local exploration around $v^{I,curr}_T$ after inspection. While the first policy enables the robot to cover its immediate surrounding environment, with subsequent detection and localization of available semantic targets, the second policy aims to travel and cover the local vicinity around $v^{I,curr}_T$ after inspection, to identify and locate distributed semantic targets. In both cases, an initial non optimized $\Tilde{\mathcal{G}}_T$ is constructed, which registers all available semantic targets within the FOV of the robot. At the end of the exploration phase, $\Tilde{\mathcal{G}}_T$ is pruned to maintain unique and high-quality semantic targets. The optimization process is performed on the basis of the consideration of spatial proximity and the segmentation characteristics. Specifically, if two $v^D$ nodes share a neighborhood, i.e. $d^{valid}_T \leq 5\unit{m}$, the node with the lower detection confidence and the lower segmented area is pruned. Moreover, if a $v^D$ node satisfies a PIP check of $v^{I,k}_T \in \{v^I_T\}$, the corresponding $v^D$ node is removed from the graph. 

Once $\mathcal{G}_T$ is obtained after optimization, $\{v^D_T\}$ are ranked based on their utility and the one with maximum utility is then chosen to be inspected. The utility of $v^{D}$ is formulated as a weighted sum of three parameters: (a) the relevance of the node, which in this work is considered to be the pixel area of the segmentation with respect to image resolution $A(v^{D}) \in \mathbb{R}^{+}$. Therefore, larger semantic targets are more likely to be inspected before smaller ones; (b) spatial proximity to other neighbouring, target nodes $N(v^{D}) \in \mathbb{R}^{+}$, i.e. nodes that are more central to other $v^D_T$ nodes are prioritized to be inspected; (c) spatial proximity to the robot $P(v^{D})  \in \mathbb{R}^{+}$. Equation~\eqref{eqn: utility_heuristic} presents the mathematical formulation of the utility heuristic function implemented to rank semantic targets for further inspection. At every necessary instant in determining the next target node to inspect,~\eqref{eqn: utility_heuristic} is evaluated at runtime and the candidate node having the maximum utility, $v^{D*}$, is then selected for inspection.


\begin{align}\label{eqn: utility_heuristic}
  v^{D*} &= \underset{v^D \in \{v^D_T\}}{\operatorname{argmax}} \, U(v^D), \\
  U(v^D) &= S_p P(v^D) + S_a A(v^D) + S_n N(v^D), \nonumber \\
  \text{where} \quad 
  P(v^D) &= \frac{1}{\| \textbf{X}_{\text{odom}[x,y,z]} - \hat{\textbf{X}}^{v^D}_{T[x,y,z]} \|}, \nonumber \\
  A(v^D) &= \frac{s^{a}_{v^D}}{I_w I_h}, \nonumber \\
  N(v^D) &= \frac{1}{\bar{d}_{v^D}}. \nonumber
\end{align}

where, $s^{a}_{v^D}$ is the area of the segmentation mask (in pixels) with $I_w,I_h$ being the width and height of the RGB image frame and $\Bar{d}_{v^D} = \frac{1}{|\{v^{D}_T\}|-1} \sum_{j=1,j\neq i}^{|\{v^{D}_T\}|-1} ||\hat{\textbf{X}}^{v^D(i)}_{T[x,y,z]} - \Hat{\textbf{X}}^{v^{D}(j)}_{T[x,y,z]}||$ is average distance of the current evaluated node $v^D(i)$ to other $v^{D}$ nodes. $S_p,S_a,S_n \in \mathbb{R}^+$ are the weights corresponding to the three modelled parameters that influence the inspection decision. Once $v^D*$ is obtained, the candidate semantic node is then subsequently communicated with the hierarchical planner (explained in Section.~\ref{sec:hpp_spp}) to provide a traversable sequence of local paths to reach the target for inspection. 

During inspection of $v^D*$, the inspection planner populates the $Level$ and $Pose$ layer graph based on the levels of inspection carried out and the respective view-pose configurations maintained at each level of inspection, according to the functionality defined in~\cite{viswanathan2023towards}. The $\textit{Feature}$ layer graph is populated by extracting segmented semantic features from the RGB image frame for every inspection view-pose.  We deploy YOLO models~\cite{yolov8_ultralytics}, specifically the Yolov8n-seg model, trained on the COCO dataset~\cite{lin2014microsoft} to segment semantic targets during exploration and a Yolov8m-seg model, trained over CarParts-seg dataset~\cite{car-seg-un1pm_dataset}, to segment semantic features of \{\textit{car},\textit{truck}\} targets during inspection.

\section{Hierarchical and Semantic Path Planning}\label{sec:hpp_spp}

\textbf{Hierarchical Path Planning:} The implemented planner performs four main operations for the given terminal nodes, ($v^{trm}_T$, $v^{trm}_L$, $v^{trm}_P$): (a) localization of the robot within $\mathcal{G}_T,~\mathcal{G}_L,~\mathcal{G}_P$ layer graphs, i.e get the $v^{I,curr}_T$, $v^{curr}_L$ and $v^{curr}_P$ nodes, (b) extract a global landmark route through $\mathcal{G}_T$ , (c)  evaluate $v^H_l$ to transition through local layers of the nodes in landmark sequence and (d) extract local traversible route based on the local layer graph and the evaluated $v^H_l$. The processes (a), (c) and (d) are recursively evaluated for each layer the planner needs to transition until the robot reaches the desired terminal nodes. To provide a traversible route over a graph, we implement Dijkstras's path planner~\cite{dijkstra1959note}.

\textbf{Remark:} \textit{To transition between high-level landmark nodes, the following conditions must be met: (a) the robot must reach node $v_L(0)$ of $v^{I,curr}_T$ and (b) plan over the local graph $\mathcal{G}_P$ to reach the nearest $v_P$ to the next landmark node. These conditions align with the $\textit{Target}$ layer traversal defined in Section~\ref{sec:layer_construction}, where LSG establishes valid connections between nearby semantic targets only if the next node is visually identified and localized during $\pi^{expl:LE}$ (i.e., the in-between region is unobstructed).}

\begin{figure}[htpb]
    \centering
    \includegraphics[width=0.8\linewidth]{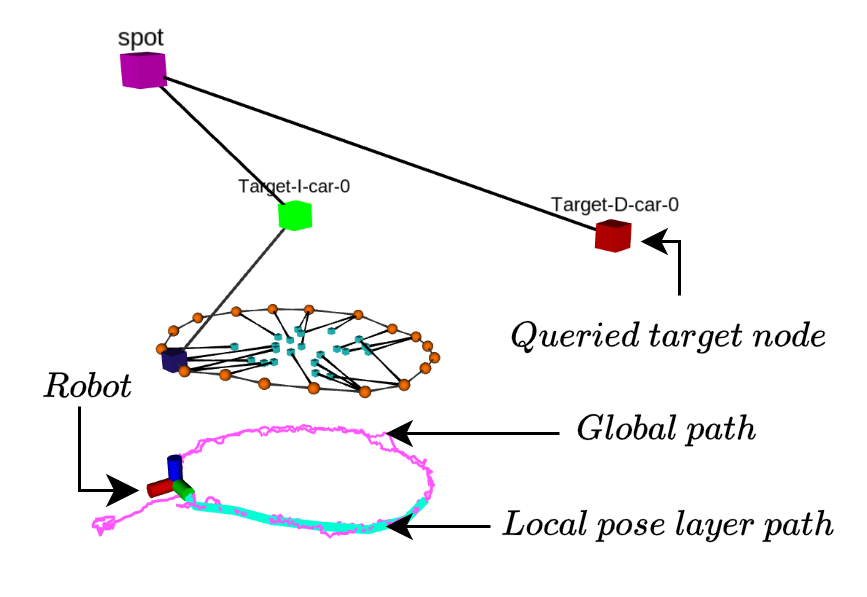}
    \caption{The response of the hierarchical planner to inspect a queried candidate inspection target, \textit{Target-D-car-0}, detected during local exploration around \textit{Target-I-car-0}.}
    \label{fig:xflie_hpp_car2_v2}
\end{figure}

From this, layer transitions are guided by $v^H_l$ nodes, evaluated based on the terminal nodes the robot must reach in the current local layer graph. For example, when navigating to the node $v^D(4)$ for inspection, i.e $v^{trm}_T=v^D(4)$, from $v^{I,curr}_T$, i.e $v^I_T(1)$ node with respect to $\mathcal{G}_T$, with current level and pose nodes $v^{curr}_L$ and $v^{curr}_P$, i.e $v_L(3)$ and $v_P(5)$ respectively, the planner must first transition out of $\mathcal{G}_P$ of $v_L(3)$ into $\mathcal{G}_L$. The layer frontier node $v^H_P = v_L(3)$, the current pose node $v_P(5)$ and the current local graph $\mathcal{G}_P$ serve as inputs to Dijkstra's path-planner. Once the robot reaches parent node $v_L(3)$, the planner transitions into the $\textit{Level}$ layer, i.e the parent layer of $v^H_P$, continuing the process of planning to the nested $\textit{Pose}$ layer of $v_L(0)$, i.e $v^H_L = v_L(0)$. Finally, the planner evaluates $v^H_P$ within $\mathcal{G}_P$ of $v_L(0)$ to transition to the next landmark node $v^I$ using nearest-neighbor search based on 3D pose data of $V_P$ and $v^I$. This process recurses until the robot reaches the terminal landmark $v^{trm}_T$ and its respective terminal nodes, if specified. Figure.~\ref{fig:xflie_hpp_car2_v2} presents the performance of the hierarchical planner when queried by FLIE to navigate towards the next candidate semantic target for further inspection. Based on the query, the local layer path generated by Dijkstra's path planner is shown as the solid \textit{cyan} colored line. 

\textbf{Semantic Path Planning:} 3D LSGs maintain an intuitive scene representation of the semantic concepts observed during the inspection mission. This makes it a powerful tool for extending the mission's lifetime by minimizing the time and effort required to process the collected inspection data. The semantic query can be represented as: $``$ Visit $v_F$ in $v_L$ of $v^I_T$ $"$, where $v_F,~v_L,~v^I_T$ denote the intended semantic labels, such as the nodes depicted in the LSG. The queried terminal nodes are then subsequently fed to the hierarchical planner for further execution. Apart from $v^{trm}_T$ and $v^{trm}_L$, which is directly parsed from the semantic-query, $v^{trm}_P$ is extracted based on the parent pose node of the queried semantic feature. 

\section{Experimental Setup and Evaluation}\label{sec:setup}

\begin{figure}[htpb]
    \centering
    \includegraphics[width=0.8\linewidth]{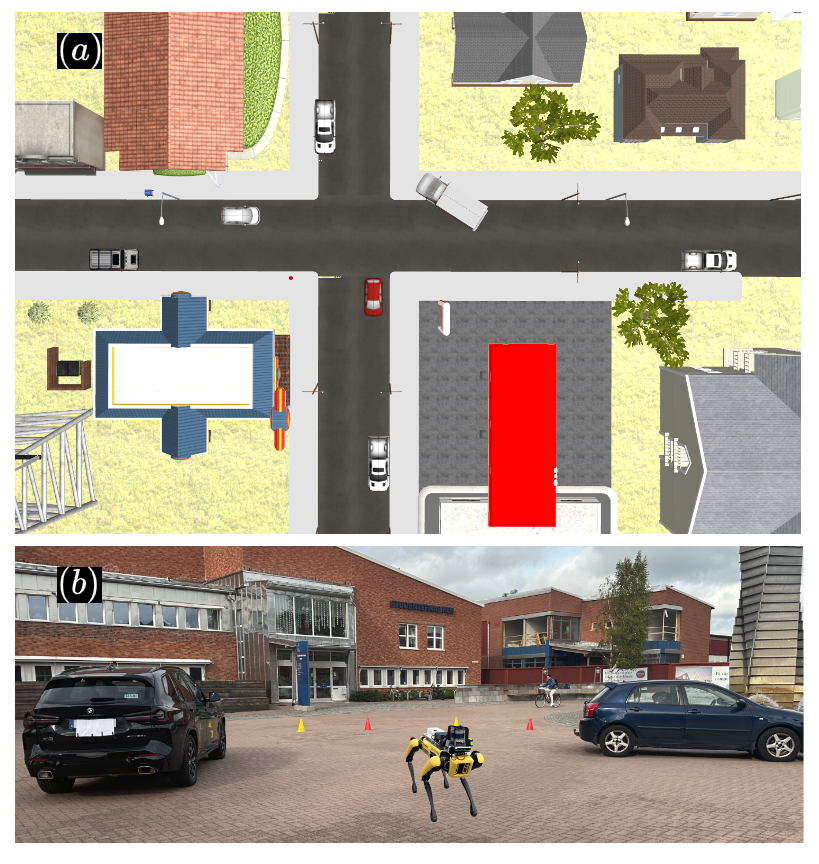}
    \caption{In (a), the small city simulation environment with seven distributed semantic targets for the evaluation of xFLIE autonomy using Gazebo and RotorS and in (b), the experimental scenario for the field evaluation of xFLIE autonomy in urban outdoor setting with two semantic targets. The scene is located within Luleå University of Technology campus grounds located in Luleå, Sweden-97187.}
    \label{fig:xflie_setup}
\end{figure}

The objective of the evaluation is to evaluate the xFLIE architecture to be able to explore, inspect and construct LSG of apriori unknown semantic targets in unknown environments. The semantic targets are considered to be \{\textit{car},\textit{truck}\} distributed around the operating environment. Additionally, we showcase a feasibility study of addressing run-time path planning queries over hierarchical representation ($HP$) against directly planning over conventional volumetric based methods ($VP$). This evaluation is made to provide insight into investigating the benefits of a hybrid path planning approach, where hierarchical path planning over scene graph could be used to generate a high-level route which is then later translated into a refined collision-safe path by the volumetric planner to navigate towards the goal. Thus, essentially transforming large-scale scene navigation into a tractable planning problem. For this, we implement $D^{*+}$\cite{karlsson2022d+}, grid-based risk-aware path planner that operates over an \textit{Octomap}\cite{wurm2010octomap} representation of the environment. We use the default values of required voxel size of 0.8 and the risk-factor of 2, as available in the repository. Figure~\ref{fig:xflie_setup} presents the outdoor urban environment simulation (refer Fig.~\ref{fig:xflie_setup}(a)) and field evaluation in outdoor campus (refer Fig.~\ref{fig:xflie_setup}(b)) environment setup in which the proposed work is validated. The proposed architecture is evaluated experimentally onboard the Boston Dynamics (BD) Spot quadruped robot equipped with Realsense D455 stereo-camera, Vectornav VN-100 Inertial Measurement Unit (IMU) and Velodyne High-res VLP16 3D LiDAR. We use Nvidia Jetson Orin embedded computational board running ROS Noetic, Ubuntu 20.04 and JetPack 5.1. For simulation, we use the RotorS simulator along with Gazebo and ROS Noetic on Ubuntu 20.04 LTS operating system with i9-13900K CPU and 128 GB RAM.

\section{Results and Discussions}\label{sec:res_des}
\begin{figure}[htbp]
    \centering
    \includegraphics[width=0.8\linewidth]{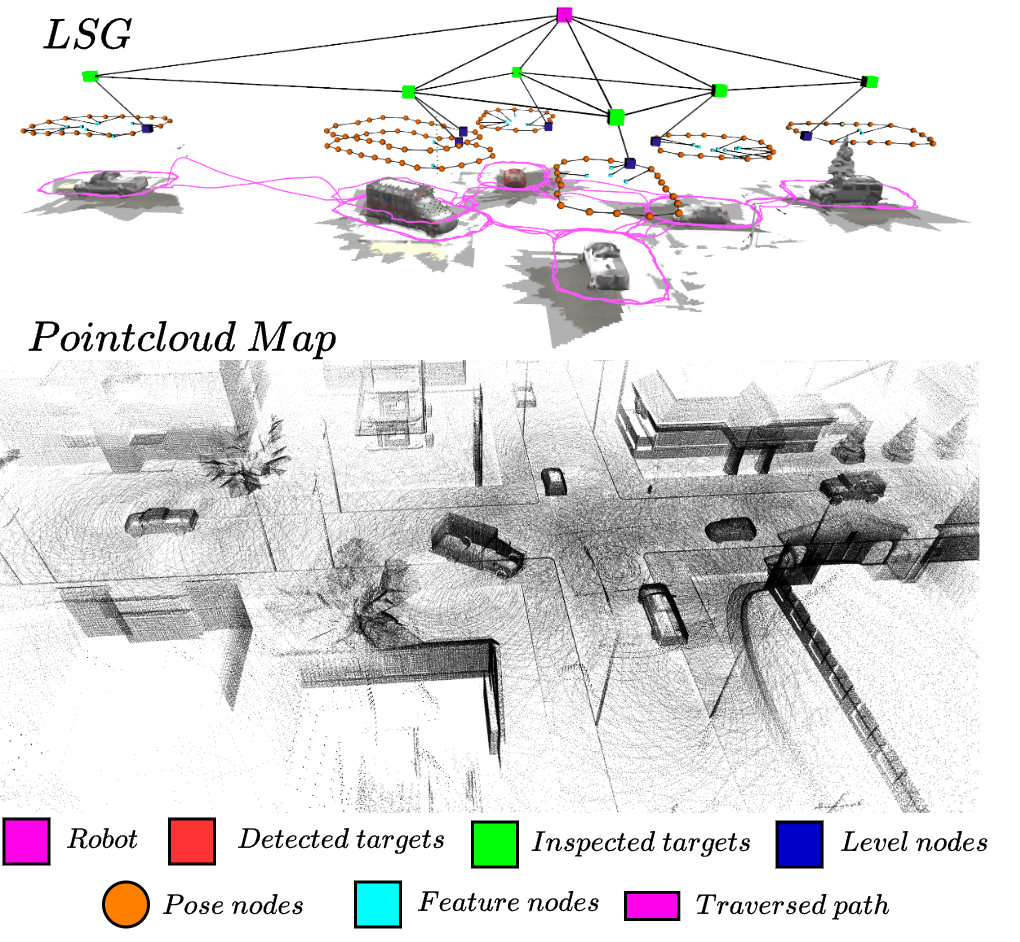}
    \caption{An overview of the constructed LSG at the end of a simulated inspection mission in urban outdoor environment. The 3D reconstructed mesh is processed through \textit{Nvblox}~\cite{millane2024nvblox} and is only used for visualization purposes.}
    \label{fig:xflie_lsg_final_sim}
\end{figure}
\begin{figure}[htbp]
    \centering
    \includegraphics[width=0.8\linewidth]{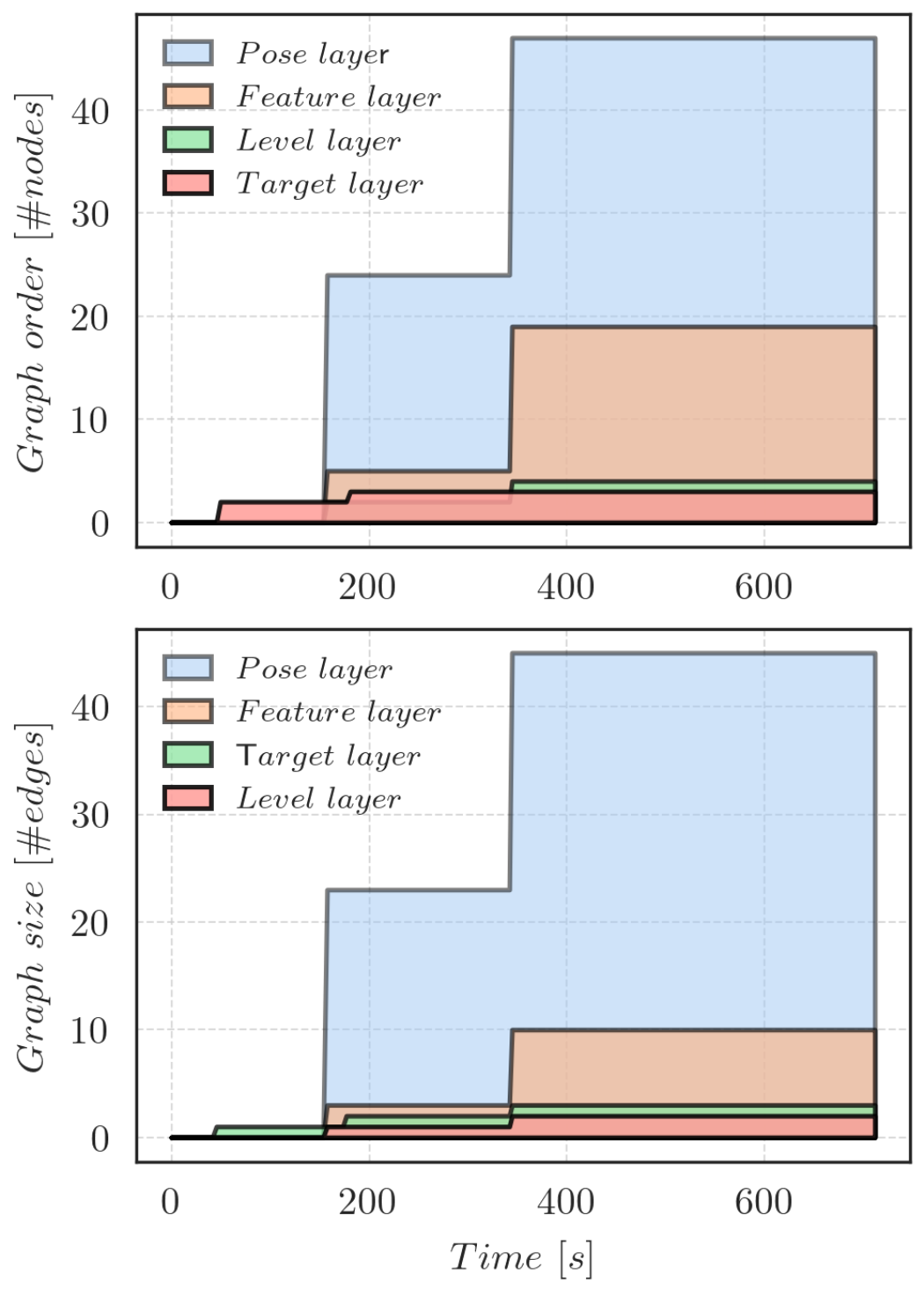}
    \caption{Time-dependent plot of the graph order property for individual layers forming the LSG throughout the simulation.}
    \label{fig:exp_lsg_info}
\end{figure}
\begin{figure}[htbp]
    \centering
    \includegraphics[width=0.8\linewidth]{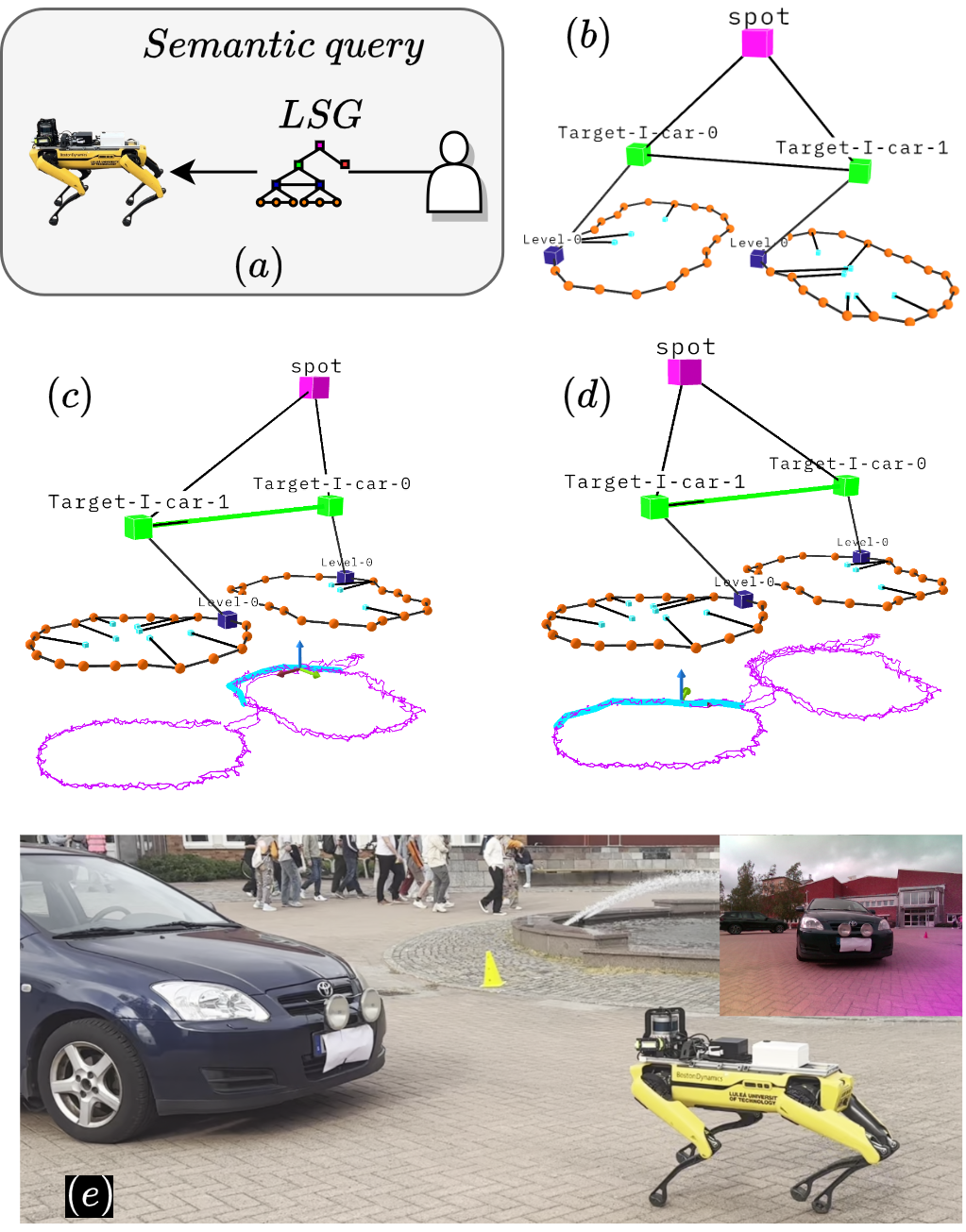}
    \caption{A collage presenting the response to a semantic query issued by a human-operator to the robot at the end of the inspection mission.}
    \label{fig:semq_car1_collage}
\end{figure}

Figure~\ref{fig:xflie_lsg_final_sim} presents the performance of the proposed xFLIE autonomy for semantic-aware inspection of apriori unknown targets-of-interest in the simulated unknown urban environment. Of the seven semantic targets distributed throughout the scene, the implemented architecture was able to successfully inspect and construct the corresponding LSG for six of them. The discrepancy is due to the performance of the onboard YOLO model to detect and segment the remaining semantic target during the exploration phase.
\begin{figure}[!htbp]
    \centering
    \includegraphics[width=0.8\linewidth]{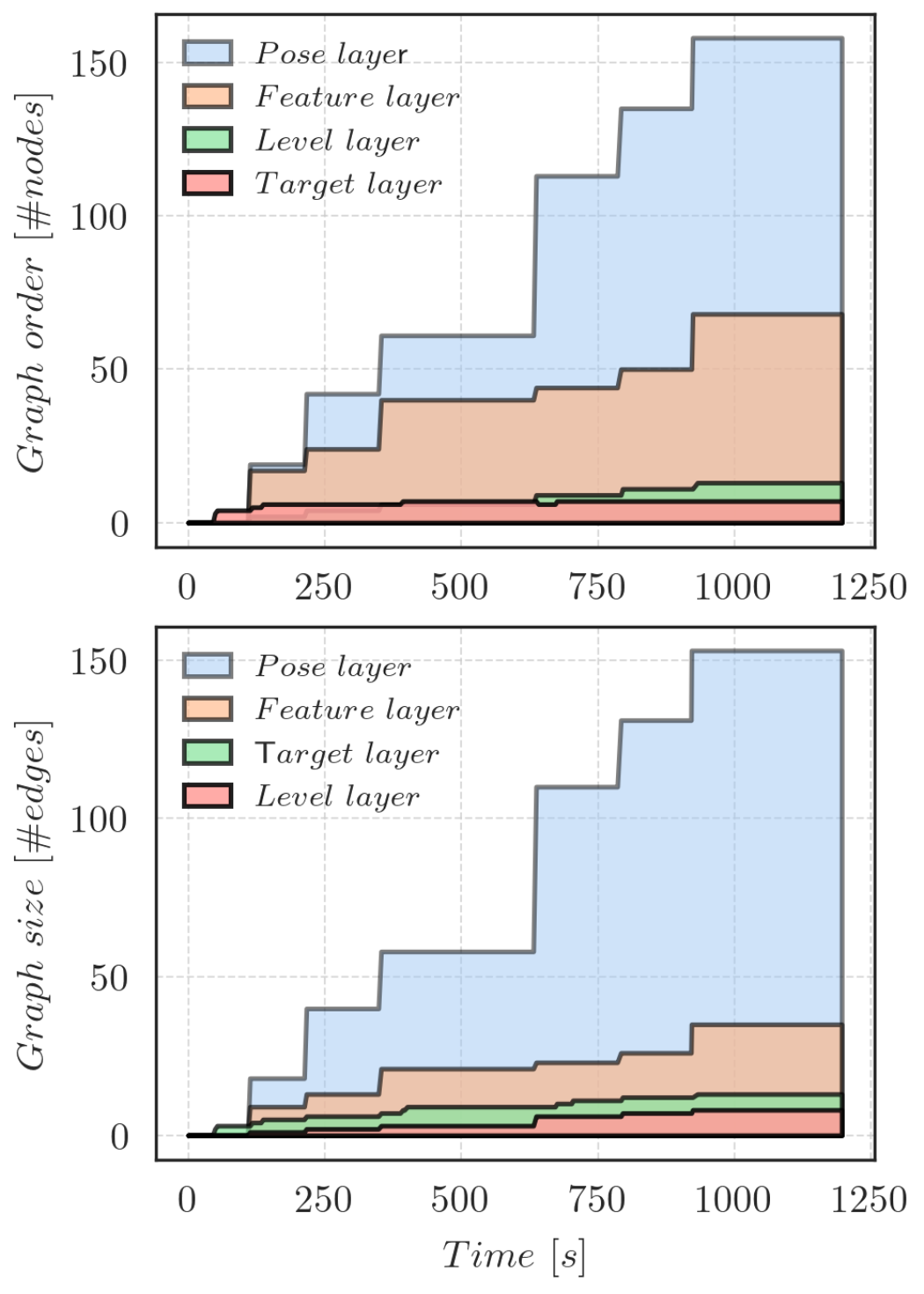}
    \caption{Time-dependent plot of the cumulative graph order property for individual layers forming the LSG throughout the simulation. Note that the color scheme for the \textit{Level} and \textit{Target} layer has been switched between the plots to improve the visibility and understanding for the reader.}
    \label{fig:sims_lsg_info}
\end{figure}
Figure.~\ref{fig:sims_lsg_info} presents a quantitative visualization into the evolution of the nested graph representations within 3DLSG during the simulated inspection mission. (\textit{On top}), the population trend in the number of nodes, i.e the graph order, registered within each layer representation is provided. (\textit{On bottom}), the growth in the number of edges, i.e.  the graph size, is plotted respective to each layer. The nature of nesting local graph representation within each node is highlighted in the evolution of \textit{Pose} and \textit{Feature} layer graphs (refer Fig.~\ref{fig:LSG_structure}). Since each local graph maintains a copy of the candidate node in the layer above as its parent, this leads to an increase in the overall number of nodes being counted for each layer. However, the constituent edges within the local layers reflect expected behavior since 3DLSG does not maintain any inter-layer connections between the nested copy and original node in the layer above. Though the results reflect the cumulative graph order for each individual local graph representation registered within the 3DLSG, the process of concatenation through the graph union operation of these local graphs would reconcile the cloned node registrations by merging duplicate nodes.  
\begin{figure}[!htbp]
    \centering
    \includegraphics[width=0.8\linewidth]{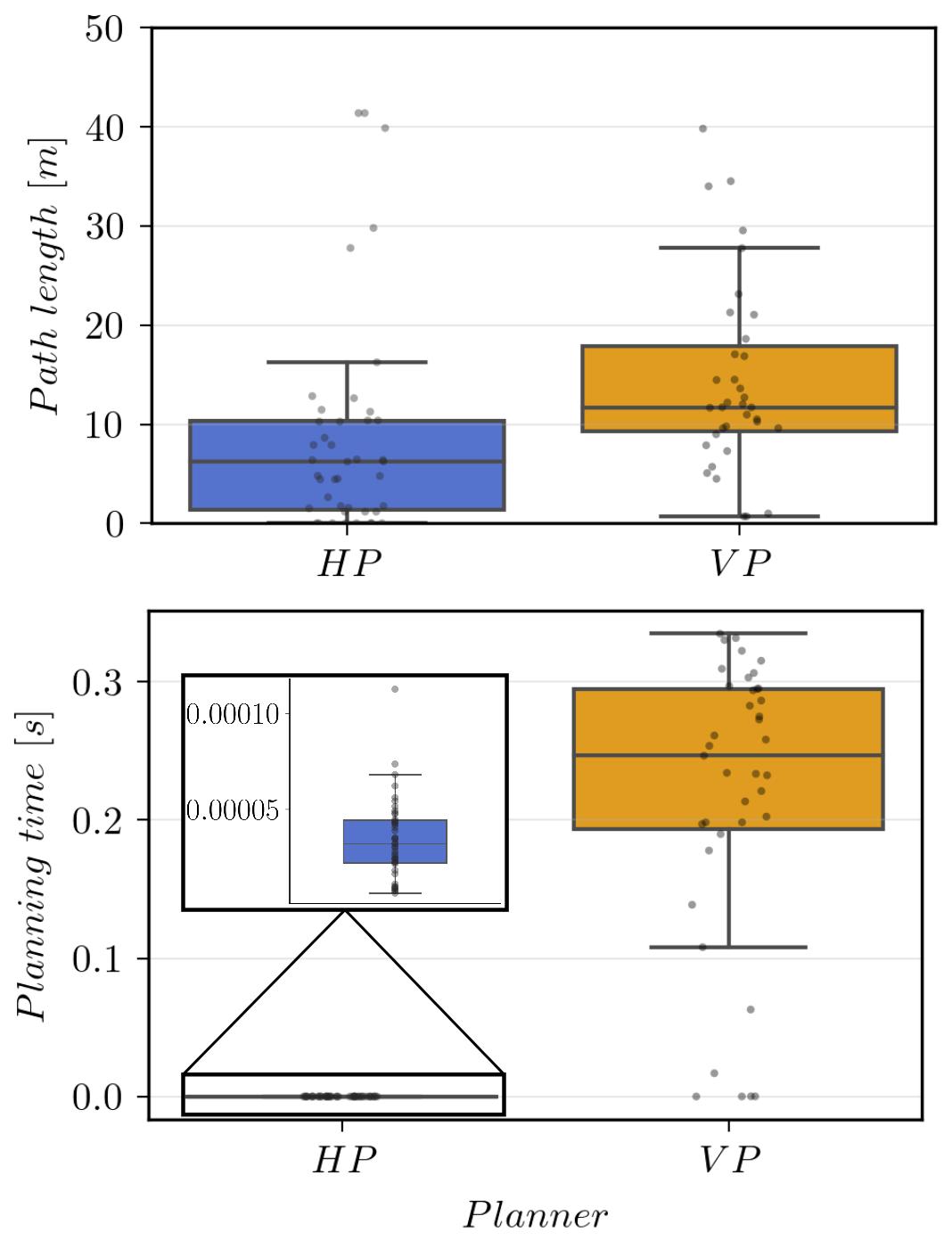}
    \caption{A run-time feasibility study recorded over the simulation run to address onboard autonomy as well as operator defined navigation queries.}
    \label{fig:plan_perf_sims}
\end{figure}
Figure.~\ref{fig:plan_perf_sims} presents performance metrics for (\textit{On top}), the length of the planned path during hierarchical path planning and the volumetric planner and (\textit{On bottom}), the corresponding time taken to generate a feasible route. Though the path planned by the $VP$ is observed to be slightly longer than the one by $HP$, $VP$ provides the safest and not necessarily the shortest route through the scene. This behavior is influenced by a combination of the risk factor value and the defined voxel size for environment representation. The recursive planning on sparse local layer graphs as the robot navigate through the scene also exhibits faster planning time to generate a traversable route. In contrast to $VP$, hierarchical path planning demonstrates orders of magnitude improvement in required planning time. In terms of real-world operations, this study underscores benefits towards initially evaluating $HP$ to generate a high-level local route and then subsequently leveraging $VP$ to generate a collision-free path. 
\begin{figure*}[!ht]
    \centering
    \includegraphics[width=0.98\linewidth]{ 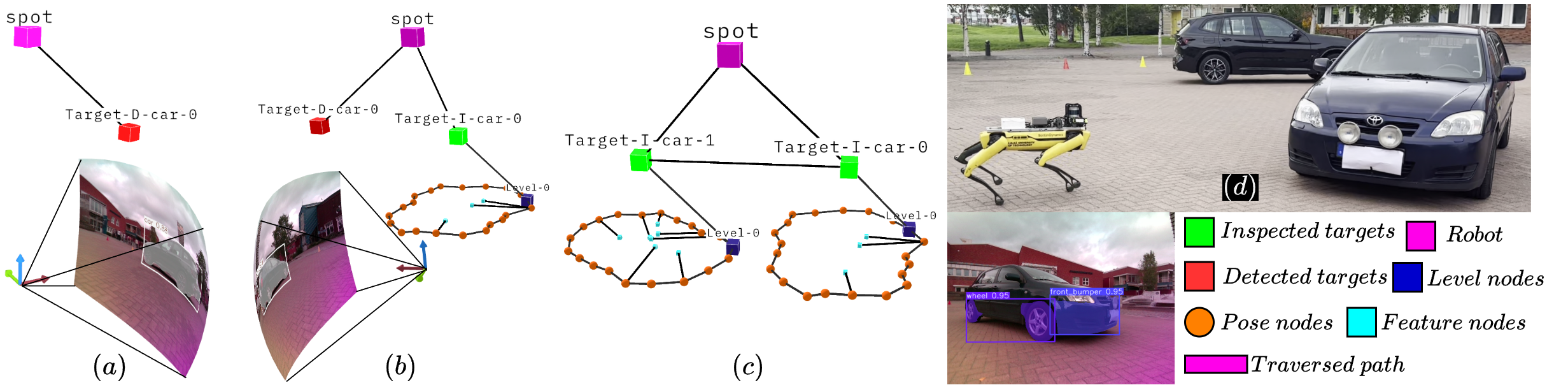}
    \caption{A collage comprising of runtime snapshots taken during field evaluation of xFLIE autonomy for a two vehicle inspection scenario.}
    \label{fig:spot_exp_twocar_21082024_run1}
\end{figure*}
Figure~\ref{fig:spot_exp_twocar_21082024_run1} presents the field evaluation of the xFLIE architecture onboard BD Spot robot for a two vehicle inspection scenario. After initialization of the mission, Fig.~\ref{fig:spot_exp_twocar_21082024_run1}(a) present the semantinc segmentation output of the YOLOv8n-seg model during exploration in addition to the corresponding registration of detected semantic targets within the $\textit{Target}$ layer of LSG. Figure.~\ref{fig:spot_exp_twocar_21082024_run1}(b) presents the detection and registration of the second car during $\pi^{expl:LE}$ policy around the first. In Fig.~\ref{fig:spot_exp_twocar_21082024_run1}(c), the final LSG composed of all layers of abstraction is visualised at the end of the inspection mission. The xFLIE architecture, overall, was able to detect, localize and inspect while constructing a LSG of then available semantic targets within the environment. The LSG is then utilized immediately for issuing additional tasks to the robots (refer Fig.~\ref{fig:semq_car1_collage} for the task of addressing semantic queries using the LSG built during the experiment). 
Figure.~\ref{fig:exp_lsg_info} captures the evolution of the local layer graphs in terms of cumulative graphs size and graph order during the outdoor urban field experiments. As the robot initially executes $\pi^{expl:360}$, the $\mathcal{G}_T$ registers observed targets. Subsequent execution of $\pi^{insp}$ populates the internal local layers of $v^{D*}$. Around $170~\unit{s}$ into the mission, inspection is terminated and the resulting graph order and size is observed within Fig.~\ref{fig:exp_lsg_info}~(\textit{top}-\textit{bottom}). Subsequent $\pi^{expl:LE}$ around the inspected target, observes, registers the remaining vehicle (\textit{Target-D-car-0}) within $\mathcal{G}_T$ and executes $\pi^{insp}$. At $\approx340~\unit{s}$, $\pi^{insp}$ is terminated and $\pi^{expl:LE}$ is executed. The increase in the corresponding population of the local layers reflect the recent update of 3DLSG. Since no new targets are observed, at $\approx420~\unit{s}$ the robot is commanded to $\textit{Return-to-base}$ by the autonomy stack. From $\approx430-710~\unit{s}$, the autonomy handles operator-defined semantic queries to visit points of interest as registered within the constructed 3DLSG. The tasked semantic queries are provided in Table.~\ref{tab:planning_results}.

Figure~\ref{fig:semq_car1_collage} presents the hierarchical and semantic path planning response to an operator-queried task: $``$ Visit \textit{front-bumper-1} in \textit{Level-0} of \textit{car-1} $"$, at the end of the inspection mission (shown in Fig.~\ref{fig:semq_car1_collage}(a)). Figure.~\ref{fig:semq_car1_collage}(b) presents the LSG constructed during the inspection mission which serves as a basis for the human-operator to issue queries. The \textit{dotted-circle} highlights the semantic \textit{front-bumper} queried to the robot (see Fig.~\ref{fig:semq_car1_collage}(b)). Figures~\ref{fig:semq_car1_collage}(c)-(d) present the response of the hierarchical planner to reach the queried $\textit{car-1}$ semantic target. The global landmark route planned through the $\textit{Target}$ layer is indicated via the solid \textit{green} line. The subsequent planned traversible paths across the local layers for transition through the landmark sequence is shown as the solid \textit{cyan} line. Finally, Fig.~\ref{fig:semq_car1_collage}(e) shows BD Spot arriving at the corresponding pose-node from where the semantic feature \textit{front-bumper-1} was observed and registered within the graph. The RGB image frame captured by the onboard Realsense D455 camera at the final position is shown on the right of Fig.~\ref{fig:semq_car1_collage}(e). 

Table.~\ref{tab:planning_results} presents a quantitative analysis of the planning performance over 3DLSG during the outdoor urban field experiments while addressing onboard autonomy ($Q_A$) as well as operator-defined semantic queries ($Q_O$). The evaluation results are presented with respect to the time, the path length and the corresponding graph considered for hierarchical path-planning. The performance metrics provided for the pose layer are with respect to the cumulative value (indicated with $^*$) registered for each planning instant over the nested $\mathcal{G}_P$. Overall, the hierarchical path planner shows consistent sub-millisecond performance in path evaluation time to address navigation queries. The latent nature of nested local layers contributes towards this performance advantage. This is due to the path planner being only exposed to a sparse local layer graph representation of the $v^I_T$ that needs to be traversed. This is recursively updated as the robot navigates through the planned global landmark route.
 Drawing insight into the evolution of the 3DLSG over time from Fig.~\ref{fig:sims_lsg_info} and presence of heterogeneous intra-layer edges (refer Sec.\ref{sec:layer_construction}), there might exist a potential computational load increase in terms of path planning when expanding to multi-task domain 3D scene graphs. This is due to the lack of consideration of explicit edge types suitable for evaluating traversability between two nodes by the graph-based path planner. Addressing the above limitation is beyond the scope of this work, therefore we identify it as a challenge to be explored in future research. 

\begin{table*}[!htbp]
\centering
\caption{Planning Performance over 3DLSG during outdoor field experiments}
\label{tab:planning_results}
\renewcommand{\arraystretch}{1.3}
\setlength{\tabcolsep}{0.1pt}
\small
\resizebox{\textwidth}{!}{%
\begin{tabular}{c P{5.8cm} *{3}{S[table-format=3.2,table-column-width=1.8cm] S[table-format=2.0,table-column-width=1.8cm] S[table-format=4.0,table-column-width=1.8cm]}}
\toprule
\multirow{2}{*}{\shortstack{Query\\Type}} & \multirow{2}{*}{\quad\quad\quad\quad\quad\quad\quad Query} & \multicolumn{3}{c}{Target layer} & \multicolumn{3}{c}{Level layer} & \multicolumn{3}{c}{Pose layer} \\
\cmidrule(lr){3-5} \cmidrule(lr){6-8} \cmidrule(lr){9-11}
 & & {Time (ms)} & {Length (m)} & {($|E|/|N|)$} & {Time (ms)} & {Length (m)} & {($|E|/|N|)$} & {Time (ms)} & {Length (m)} & {($|E|/|N|)$} \\
\midrule
$Q_A$ & \quad\quad\quad\quad\quad \textit{Target-D-car-0} 
                              & Nil    & Nil    & \multicolumn{1}{c}{1/2}     
                              & Nil    & Nil    & \multicolumn{1}{c}{1/2}      
                              & \multicolumn{1}{c}{0.178}  & \multicolumn{1}{c}{12.541} & \multicolumn{1}{c}{23/24} \\
$Q_A$ & \quad\quad\quad\quad\quad\textit{Return to Base}   
                              & \multicolumn{1}{c}{0.0486} & \multicolumn{1}{c}{9.798}  & \multicolumn{1}{c}{3/3}      
                              & Nil    & Nil    & \multicolumn{1}{c}{2/4}      
                              & \multicolumn{1}{c}{0.120}  & \multicolumn{1}{c}{12.541} & \multicolumn{1}{c}{22/23} \\
$Q_O$ & $Visit~\textit{front-bumper-1}~in~\textit{Level-0}~of~\textit{car-1}$ 
                              & \multicolumn{1}{c}{0.05302} & \multicolumn{1}{c}{9.798}  & \multicolumn{1}{c}{3/3}      
                              & Nil    & Nil    & \multicolumn{1}{c}{2/4\textsuperscript{*}}      
                              & \multicolumn{1}{c}{0.2018\textsuperscript{*}} 
                              & \multicolumn{1}{c}{19.499\textsuperscript{*}} 
                              & \multicolumn{1}{c}{45/47\textsuperscript{*}} \\
$Q_O$ & $Visit~\textit{front-right-door-1}~in~\textit{Level-0}~of~\textit{car-0}$
                              & \multicolumn{1}{c}{0.0542}  & \multicolumn{1}{c}{9.798}  & \multicolumn{1}{c}{3/3}      
                              & Nil    & Nil    & \multicolumn{1}{c}{2/4\textsuperscript{*}} 
                              & \multicolumn{1}{c}{0.1602}  
                              & \multicolumn{1}{c}{13.14} 
                              & \multicolumn{1}{c}{23/24} \\
$Q_O$ & $Visit~\textit{tailgate-1}~in~\textit{Level-0}~of~\textit{car-1}$
                              & \multicolumn{1}{c}{0.0593}  & \multicolumn{1}{c}{9.798}  & \multicolumn{1}{c}{3/3}     
                              & \multicolumn{1}{c}{Nil}    & \multicolumn{1}{c}{Nil}    & \multicolumn{1}{c}{2/4\textsuperscript{*}} 
                              & \multicolumn{1}{c}{0.1754\textsuperscript{*}} 
                              & \multicolumn{1}{c}{12.579\textsuperscript{*}}     
                              & \multicolumn{1}{c}{45/47\textsuperscript{*}} \\
\bottomrule
\end{tabular}%
}
\medskip
\footnotesize\\
$*$ calculated cumulative value, $|E|$= number of edges, $|N|$= number of nodes 
\end{table*}

\section{Conclusions}\label{sec:conclusions}

In this article, we introduced xFLIE, a pioneering solution that integrates the high-level FLIE mission planner with the construction of LSG, an actionable hierarchical scene representation for semantic-aware inspection in unknown environments. As an alternative to conventional volumetric-map based approaches, we demonstrate through simulations and field deployments on BD Spot how intuitive representations could enhance situational awareness for human operators as well as for the onboard autonomy, especially during large-scale inspection missions. Moreover, we successfully validate the concept of extending the lifetime of the inspection mission by addressing operator-defined semantic-queries. In view of future works, consideration of optimal volumetric exploration polices can help the robot to cover and detect inspection targets in larger environments effectively.
\bibliography{bib}

\end{document}